%% file: main.tex
\documentclass[final]{cvpr}

\usepackage{times}
\usepackage{epsfig}
\usepackage{graphicx}
\usepackage{amsmath}
\usepackage{amssymb}

\usepackage[pagebackref=true,breaklinks=true,colorlinks,bookmarks=false]{hyperref}

\usepackage{xparse}

\usepackage{color}

\usepackage[utf8]{inputenc}
\usepackage[T1]{fontenc}
\usepackage{newtxtext}
\usepackage{booktabs}       
\usepackage{microtype}      

\usepackage{etoolbox}
\usepackage[enable]{easy-todo}

\usepackage{graphicx}
\usepackage{multirow}

\usepackage{subfig}

\usepackage{algorithmic,algorithm}

\def\cvprPaperID{****} 
\def\confYear{CVPR 2021}

\input defs.tex

\begin{document}

\title{Generating Structured Adversarial Attacks Using Frank-Wolfe Method}

\author{Ehsan Kazemi\\
University of Central Florida\\
{\tt\small ehsan\_kazemy@knights.ucf.edu}
\and
Thomas Kerdreux\\
Centre de Recherche INRIA de Paris\\
{\tt\small thomaskerdreux@gmail.com}
\and
Liquang Wang\\
University of Central Florida\\
{\tt\small lwang@cs.ucf.edu}
}

\maketitle

\begin{abstract}
White box adversarial perturbations are generated via iterative optimization algorithms most often by minimizing an adversarial loss on a $\ell_p$ neighborhood of the original image, the so-called distortion set. Constraining the adversarial search with different norms results in disparately structured adversarial examples. Here we explore several distortion sets with structure-enhancing algorithms. These new structures for adversarial examples might provide challenges for provable and empirical robust mechanisms. Because adversarial robustness is still an empirical field, defense mechanisms should also reasonably be evaluated against differently structured attacks. Besides, these structured adversarial perturbations may allow for larger distortions size than their $\ell_p$ counter-part while remaining imperceptible or perceptible as natural distortions of the image. We will demonstrate in this work that the proposed structured adversarial examples can significantly bring down the classification accuracy of adversarialy trained classifiers while showing low $\ell_2$ distortion rate. For instance, on ImagNet dataset the structured attacks drop the accuracy of adversarial model to near zero with only 50\% of $\ell_2$ distortion generated using white-box attacks like PGD. As a byproduct, our finding on structured adversarial examples can be used for adversarial regularization of models to make models more robust or improve their generalization performance on datasets which are structurally different.
\end{abstract}

\section{Introduction}\label{sec:introduction}
Adversarial examples are inputs to machine learning classifiers designed to cause the model to misclassify the input images. These samples are searched in the vicinity of some samples in the test set,  and typically in their norm-ball neighborhoods, the so-called \textit{distortion set}. When replacing every test set samples with their corresponding adversarial examples, the accuracy of standardly trained classifiers drops to zero in the inverse correlation with the considered norm-ball radius. Thus, the lack of robustness of classifiers to adversarial samples challenges the security of some real-world systems and pose questions regarding the generalizing properties of neural classifiers \cite{schmidt2018adversarially,stutz2019disentangling}.

Thus far, there have been some successful studies on defense strategies against adversarial examples, though most of the attack and defense mechanisms considered $\ell_p$ neighborhoods. In particular, existing approaches for learning adversarially robust networks include methods which are both empirically robust via adversarial training \cite{goodfellow2014explaining, kurakin2016adversarial, madry2017towards} and also certifiably robust with certified bounds \cite{wong2017provable, raghunathan2018certified, zhang2019towards} and randomized smoothing \cite{cohen2019certified, yang2020randomized}. Recently, there were some studies  which outlined the inherent limitations of the $\ell_p$ balls \cite{sharif2018suitability,sen2019should}. While some recent papers \cite{xu2018structured,wong2019wasserstein} pointed out the benefits of other families of distortions sets, many classical norm families remained mostly unexplored in the adversarial setting.
In this work we consider white-box adversarial attacks on neural networks. In the white-box framework, the model and the in place defenses are known to the attacker. Adversarial examples in this framework are typically crafted using optimization algorithms aiming to minimize constrained adversarial losses. In the black-box attacks the attacker can only make queries and observe the response of the model. 

\begin{figure}[h!]
    \centering
    \subfloat[FGSM]{{\includegraphics[scale=0.3]{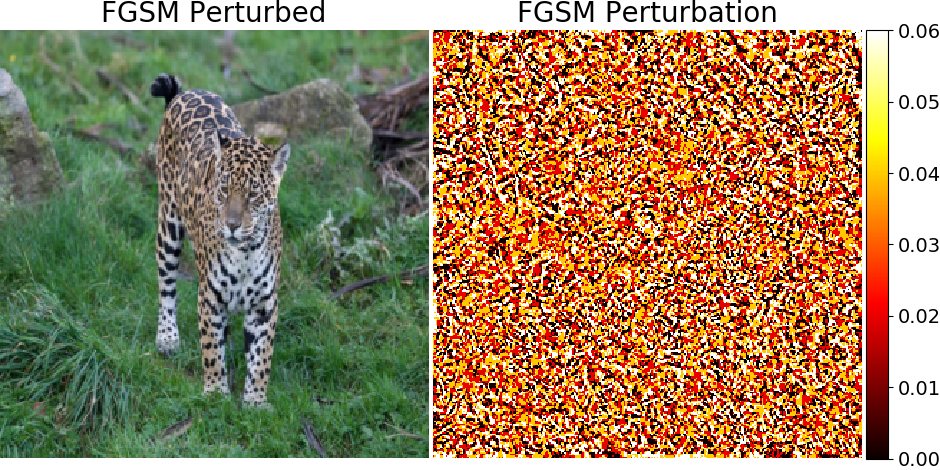}}}\hfill
    \centering
    \subfloat[FW with nuclear ball (FWnucl )]{{\includegraphics[scale=0.3]{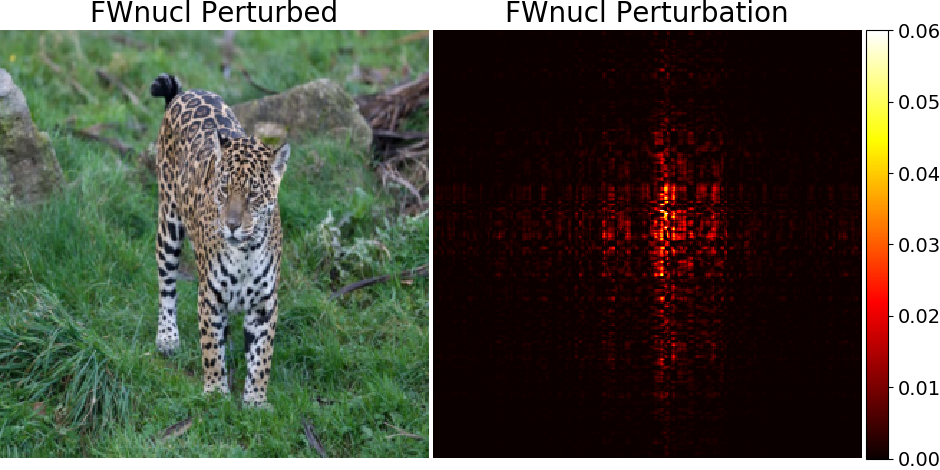}}}
    \caption{The images correspond to two types of targeted attacks. Projected Gradient Descent (PGD) solve \eqref{eq:opt_problem_adversarial} constrained by a $\ell_{\infty}$ ball while FWnucl solves \eqref{eq:opt_problem_adversarial} constrained with a nuclear ball. The type of adversarial perturbations differs significantly in structure.}
    \label{fig:example_diff_distortion}
\end{figure}

Although norms are equivalent in the image finite-dimensional space, the type of norm-ball influences the structure of the optimization algorithm iterations and the (local) minima to which they converge. As the studies on the robustness of neural model still remained empirical, it is hence necessary to explore the effect of particular structures in adversarial perturbation besides $\ell_p$ balls. For instance, Figure \ref{fig:example_diff_distortion} shows that the perturbation generated by the proposed attack (FWnucl) is more structured and targeted to the main objective in the image. 
Relatedly, important security concerns can be raised when some empirical defense mechanisms are vulnerable to certain pattern in the adversarial examples. Thus, providing a catalog of many structured attacks would cause rapid development of robust machine learning algorithm due to an arms race between attack and defense mechanisms and can greatly expand the scope of adversarial defenses to new contexts. For instance in \cite{carlini2019evaluating}, it is shown many defense mechanisms can be broken by stronger attacks, while exhibiting robustness to the weaker attacks. Thus, finding more diverse attacks is important for evaluating defense strategies. In addition, as adversarial training uses attack methods as a form of regularization for training neural networks, the training process can be performed on the newly proposed adversaries to robustify models against discovered structured semantic perturbations. These sorts of training process can better flatten the curvature of decision boundaries which can be potentially an important parameter to improve generalization performance in non-adversarial setting \cite{keskar2016large}. 

Regarding with generating the white-box adversarial samples, the radius of the convex balls is often considered sufficiently small to ensure that the added perturbations to the original samples are imperceptible. This imperceptibility requirement is pervasive in the literature, although it is not the only studied regime for adversarial examples \cite{gilmer2018motivating}. Arguably, the imperceptibility of the distortion does not play a crucial role in crafting adversaries, in particular when the ideal level of perturbation is aligned with the human perception in the sense that the perturbed image is labeled as the original image for a human observer.
In fact, the imperceptible deformation regime of non-robust classifiers has received much attention because it highlights the gap between human perception and the processing done by machine learning systems to classify non-perceptible class of perturbed samples \cite{gilmer2018motivating}.

In this work, we do not limit ourselves to the imperceptible regime of perturbation. Instead, we explore adversarial examples' structure leading to possibly perceptible deformations that would yet be considered as \textit{non-suspicious} alteration of the image. In particular, we consider the trace norm ball (the nuclear ball), which is the convex relaxation of rank-1 matrices. Qualitatively, adding perturbation in this distortion set leads to blurring effects on the original image. This blurring effect could be further localized in a controlled way to specific semantic areas in the image by considering the group-nuclear ball distortion set, where the groups are defined on the specific semantic area of interest.

In the sequel, for the sake of simplicity of the presentation we focus on untargeted adversarial examples. Our approach is to use an auxiliary optimization problem to craft the adversarial perturbations. The optimization problem to generate untargeted adversarial attack for the original sample $x^{ori}$ is formulated by 
\begin{align}\label{eq:opt_problem_adversarial}
\begin{split}
\mbox{minimize} & \mathcal{L}(x) = -L(f(x), y)\\
\mbox{subject to} &  ~ ||x-x^{ori}|| \leq \epsilon
\end{split}
\end{align}
where $L$ is an adversarial loss (e.g., cross entropy loss), $f$ is the neural classifier and $y$ is the label of the original sample $x^{ori}$. In this formulation, $\epsilon$ constrains the perturbation magnitude in particular norms. 
\paragraph{Related Work.}
Several recent research studies question the underlying reason for considering $\ell_p$ neighborhood as distortion sets and propose alternative adversarial models. For instance, \cite{sharif2018suitability} suggest that $\ell_p$ norms are neither the right metric for perceptually nor even content-preserving adversarial examples. In \cite{sen2019should} a behavioral study is conducted which shows that $\ell_p$ norms and some other metrics do not align with the human perception.

There are some recent works which consider adversarial perturbations beyond the $\ell_p$ distortion sets. In \cite{engstrom2017rotation} it is shown that simple rotation and translation can create efficient adversarial examples. \cite{xu2018structured} consider group-lasso distortion sets which are optimized based on methods like ADMM. \cite{liu2018beyond} generate adversarial examples based on the geometry and physical rendering of the image. They notably suggest that \textit{large pixel perturbations can be realistic if the perturbation is conducted in the physical parameter space (e.g., lighting)}.
\cite{wong2019wasserstein} recently argue that robustness to Wasserstein perturbations of the original image is essentially an invariant that should typically exist in classifiers. Recently, \cite{wong2020learning} investigate learning perturbation sets without optimization based approaches and via applying conditional generative models. 

There exists some methods which solve the adversarial optimization problem on specific subspaces, which might lead to specifically structured adversarial examples. While a random subspace \cite{Subspace19} does not necessarily induce specific perturbation structure, projection on low-frequency domain \cite{guo2018low} or onto the subspace generated by the top few singular vectors of the image \cite[\S 3.4.]{yang2019me} will induce structured adversarial examples. These approaches are leveraged to reduce the search space of adversarial perturbation for more efficient computational complexity. Finally, one can consider the problem of adversarial attack generation as an image processing task. A recent trend to various types of such algorithms are for instance conditional or unconditional generative models, style transfer algorithms, or image translation algorithms \cite{reed2016learning,gatys2017controlling,risser2017stable,lu2017decoder}.

In this paper we particularly apply Frank-Wolfe methods to solve adversarial optimization problem. These algorithms have shown a recent revival in constrained optimization problems for machine learning, where their success is notably due to their low cost computational cost per iterations \cite{jaggi2013revisiting}. It is known that Frank-Wolfe method exhibits linear convergence on polytopes \cite{guelat1986some,garber2013linearly,garber2013playing,
lacoste2013affine,FW-converge2015}, on strongly convex set \cite{levitin1966constrained,demyanov1970,dunn1979rates,garber2015faster} or uniformly convex sets \cite{kerdreux2020uniformly}. Frank-Wolfe algorithm has been extensively studied in convex setting for large scale nuclear norm regularization \cite{jaggi2010simple,lee2010practical,shalev2011large,harchaoui2012large,dudik2012lifted,allen2017linear,garber2018fast}. Furthermore, many variations of Frank-Wolfe method exist \cite{freund2017extended,cheung2017projection} that leverage the facial properties to preserve structured solutions for non-polytop or strongly convex domains. A closer approach to this work is \cite{chen2018frank}, where the authors apply zero-order Frank-Wolfe algorithm for solving adversarial problems in the black-box setting.

\paragraph{Contribution.} Currently the defense techniques and in particular the mechanisms which provide theoretical guaranties are designed for non-structured norms while structured norms are largely overlooked in literature. This shortcoming may render previous defense algorithms less appealing when exposed to structured adversaries.  We study some families of structured norms in the adversarial example setting. This is a pretense to more generally motivate the relevance of structured attacks (\textit{i.e.} besides the $\ell_p$ distortion set), that are largely unexplored.
It is also a versatile approach to produce specific modification of the adversarial images, like (local) blurriness. We demonstrate in the experiments that the proposed structured adversaries generates samples which target the important parts of image resulting in a lower number of perturbed elements from the original image, and therefore provide a lower perturbation magnitude which make them undetectable (see Figure \ref{fig:example_diff_distortion}). To the best of our knowledge it is the first time an optimization method is exploited to generate adversarial attacks by imposing blurriness to the target images and this work could be considered as a first step to develop more sophisticated approaches in future. Currently, the commonly-used packages for crafting adversarial samples, e.g., Foolbox \cite{rauber2017foolbox} apply spatial filters aiming to craft adversaries via blurring. We also demonstrate an algorithm for the localized perturbations (blurriness) of the region of interest in the image using group norms. 

\section{Structured Distortion Sets}\label{sec:structured_distortion_sets}

Here we describe some structured families of norms that to the best of our knowledge have not so far been explored in the context of adversarial attacks. To be more specific, we generate some specific structured perturbations by solving the adversarial problem \eqref{eq:opt_problem_adversarial}, which  provides the potential attacker a framework to derive adversarial alternation of the original test samples. In sequel, we set the trace norm ball as the distortion set  and design a framework to solve the optimization problem \eqref{eq:opt_problem_adversarial} based on conditional gradient algorithms. 
In the conditional gradient algorithm, in each iteration a Linear Minimization Oracle (LMO) is solved. More technically, for a direction $d$ and a convex set $\mathcal{C}$, the LMO problem is defined as
\BEQ\label{eq:general_definition_LMO}
\text{LMO}_{\mathcal{C}}(d) \in\underset{v\in\mathcal{C}}{\argmin } ~ d^T v.
\EEQ
The iterations of conditional gradient algorithms are then constructed as a (sparse) convex combination of the solutions to \eqref{eq:general_definition_LMO}. These solution points can always be chosen as the vertices of $\mathcal{C}$. Hence, the specific structure of the solutions of the LMO is applied in the early iterations of  optimization problem. In the following section we provide the mathematical formulation of optimization problem.

\subsection{Low-rank perturbation}\label{ssec:low_rank_perturbation}
We let $||\cdot||_{S1}$ denote the nuclear norm which is the sum of the matrix singular value, a.k.a. the trace norm or the $1$-Schatten norm. The nuclear norm has been classically used to find low-rank solutions of convex optimization problems \cite{fazel2001rank,candes2009exact} such as matrix completion. 
Here, we propose to simply consider nuclear balls as distortion sets when searching for adversarial examples in problem \eqref{eq:opt_problem_adversarial}. We later explain the various potential benefits of using this structural distortion set. To our knowledge, the low-rank structure is leveraged in different aspects of some defense techniques \cite{yang2019me} but it has never been acquired to craft adversarial attacks. 
As an empirical defense mechanism, \cite{langeberg2019effect} add a penalization in the training loss to enhance the low-rank structure of the convolutional layer filters. \cite{yang2019me} notably propose a pre-processing of the classifier outputs, which randomly removes some input pixels and further reconstructs it via matrix completion for denoising purposes. 

\begin{figure}
    \centering
    \captionsetup[subfloat]{labelformat=empty}
    \subfloat[Original]{\includegraphics[width=0.19\linewidth]{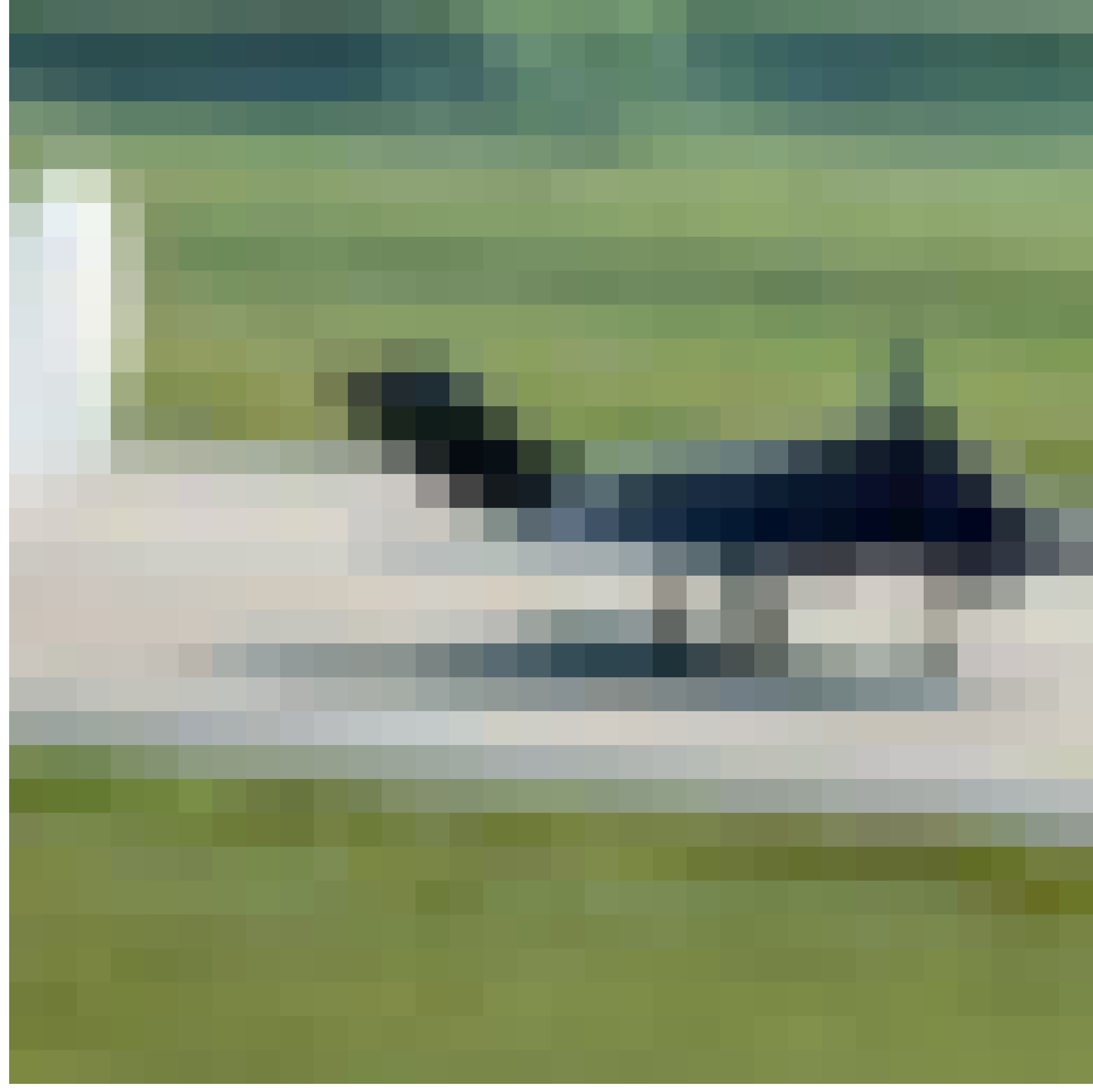}}
    \subfloat[$\epsilon_{S1}=5$]{\includegraphics[width=0.19\linewidth]{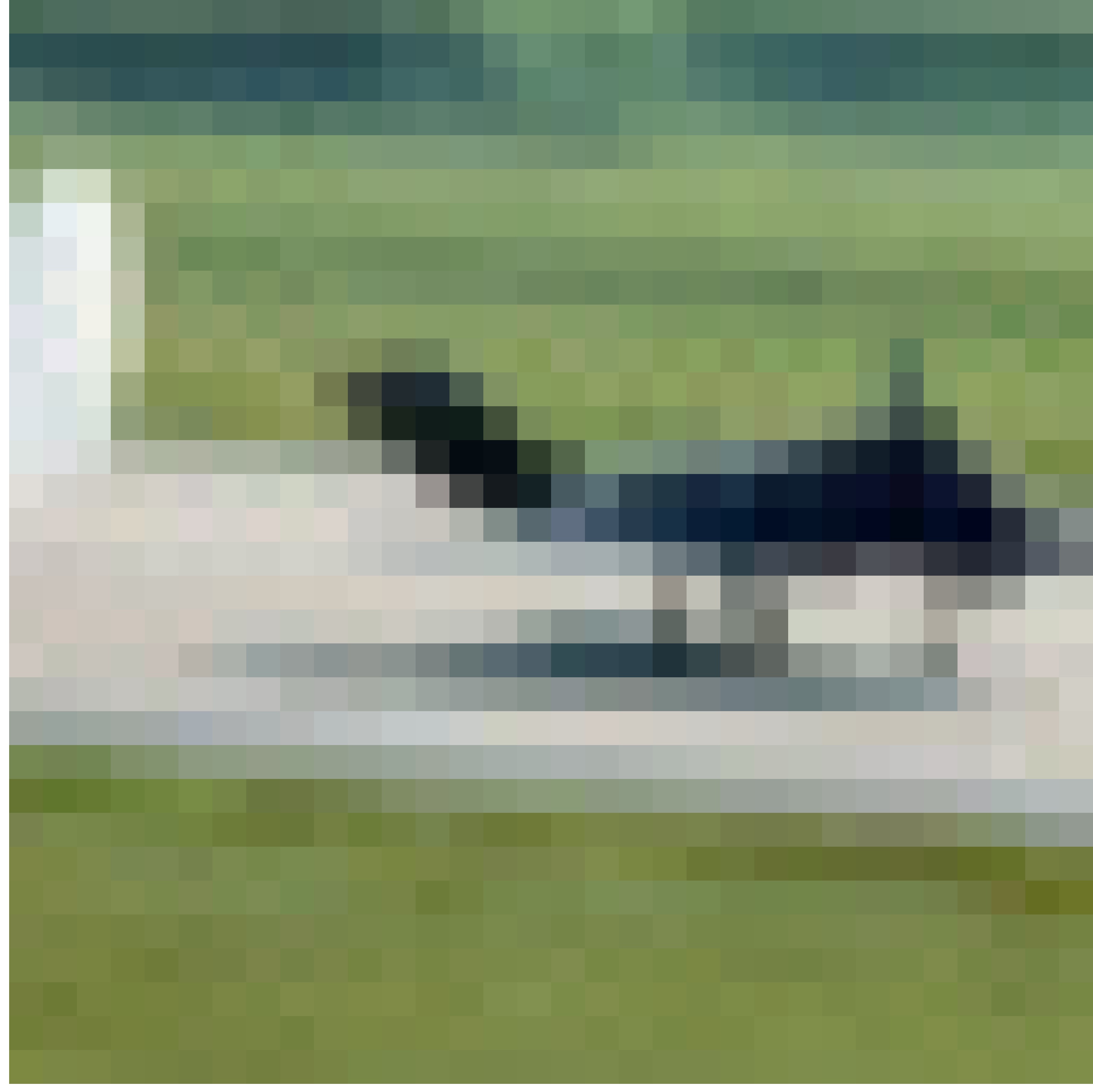}}
    \subfloat[$\epsilon_{S1}=10$]{\includegraphics[width=0.19\linewidth]{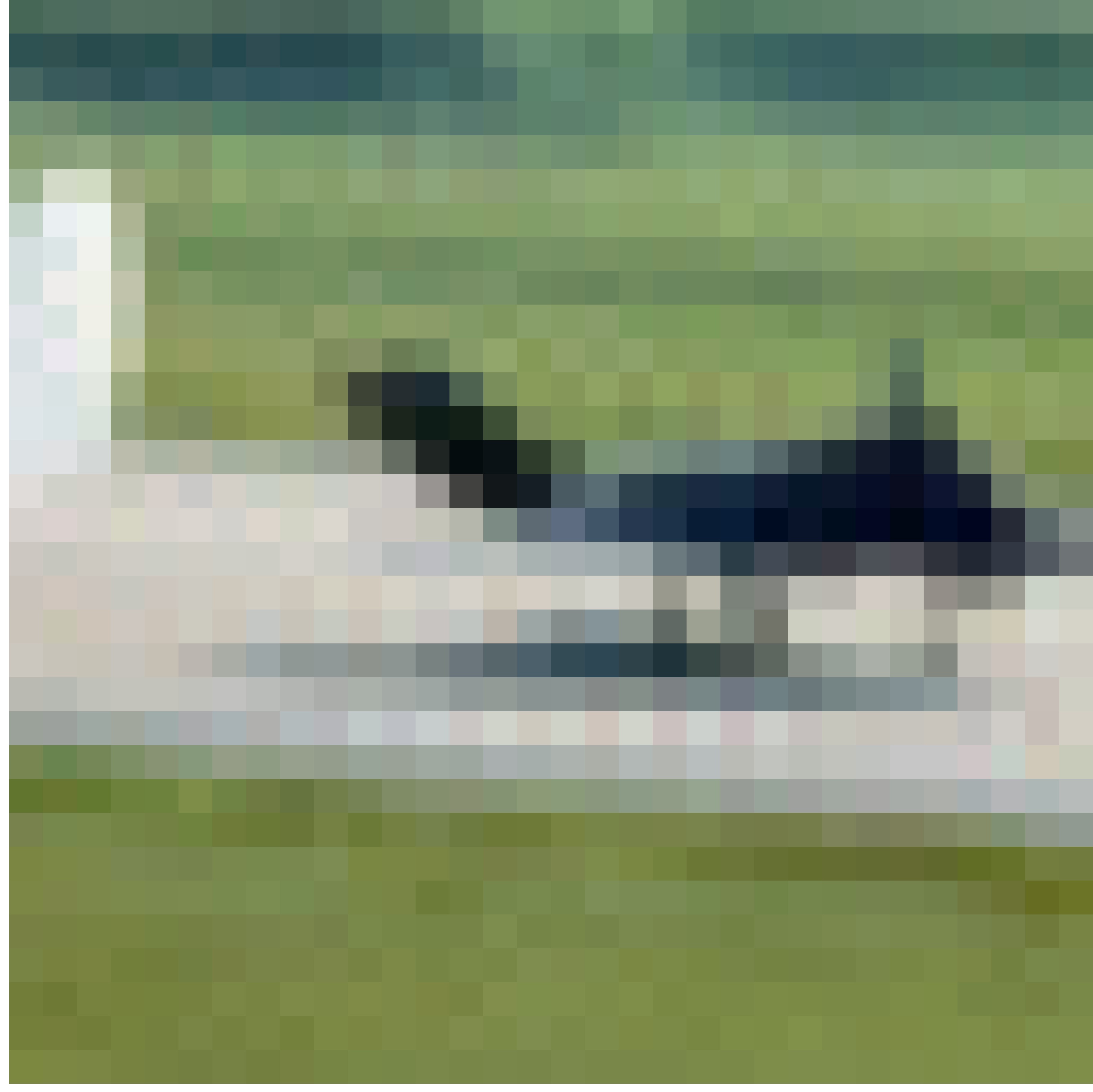}}
    \subfloat[$\epsilon_{S1}=20$]{\includegraphics[width=0.19\linewidth]{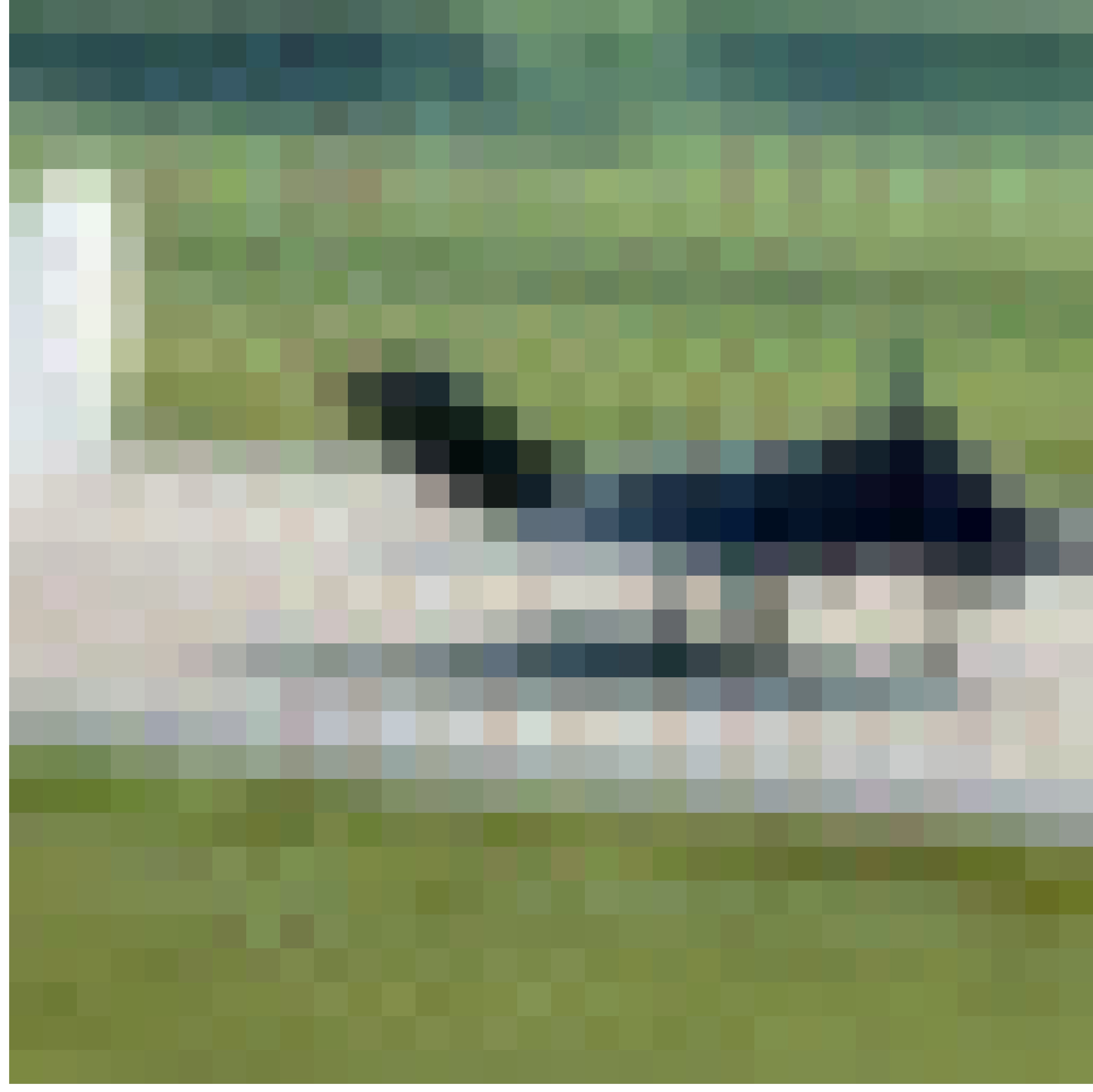}}
    \subfloat[$\epsilon_{S1}=30$]{\includegraphics[width=0.19\linewidth]{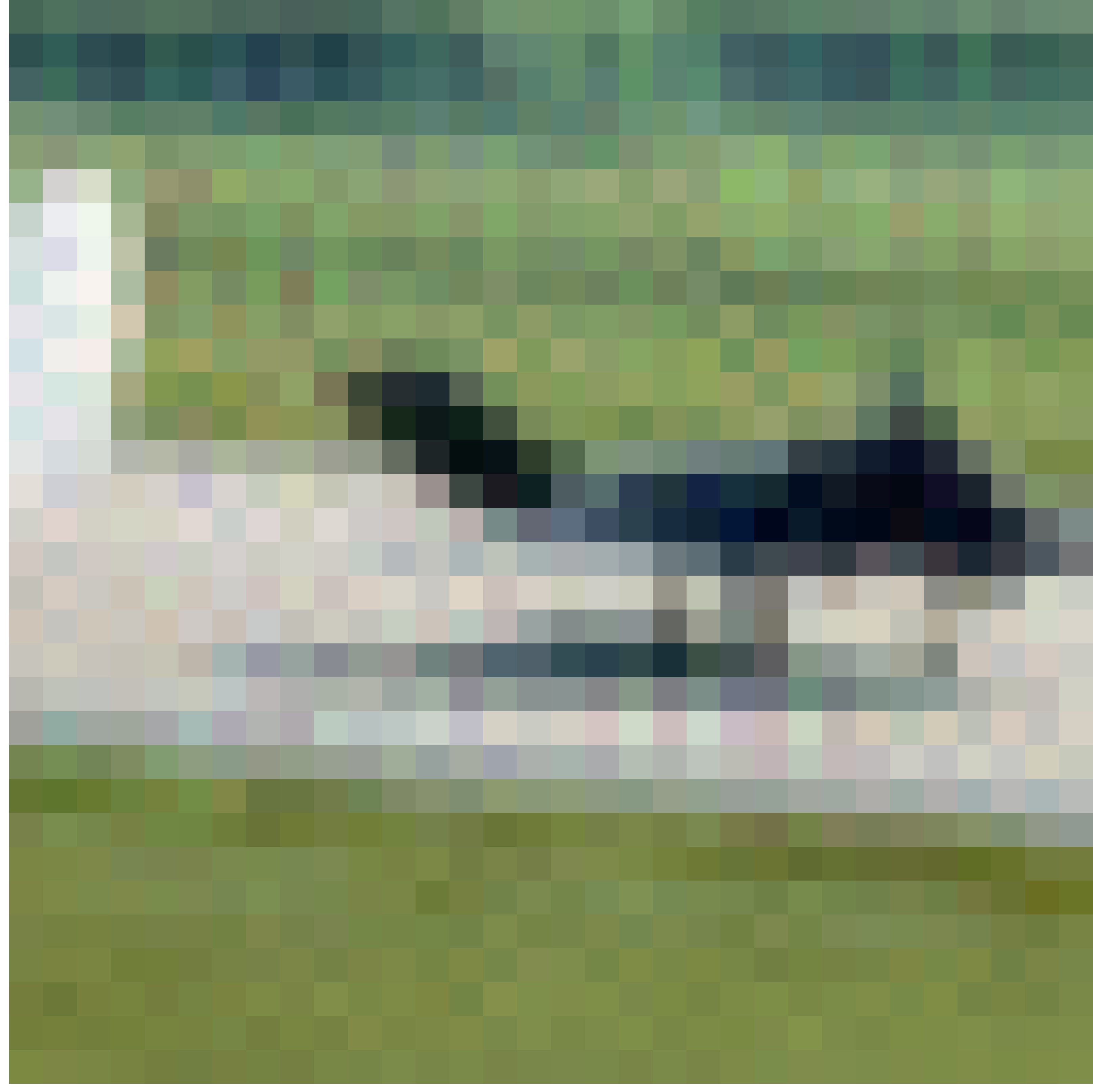}}
    
    \caption{For a test image of CIFAR-10, we computed the various adversarial examples stemming from solving \eqref{eq:opt_problem_adversarial} on the nuclear ball with Frank-Wolfe algorithm. From left to right: original image, adversarial example with a nuclear radius of $\epsilon_{S1}=5, 10, 20, 30$. Note that the adversarial examples are already miss-classified with $\epsilon_{S1}=3$; here we increase the radius purposely to observe the perturbation on the initial image.}
    \label{fig:nuclear_blur}
\end{figure}

More formally, with nuclear ball as a distortion set, the adversarial optimization problem is defined as 
\BEQ\label{eq:low_rank_adversarial}
\underset{||x-x^{ori}||_{S1} \leq \epsilon}{\argmin } \mathcal{L}(x) = -L(f(x), y).
\EEQ
This formulation is a particular example of the family of $p$-Schatten norms $||\cdot||_{Sp}$, i.e., the $p$-norm of the singular value vector with $p=1$. These structured norms lead to differently structured adversarial examples.  Given the lack of explicit mathematical translation across norms, this adversaries may end up to defeat certified approaches in terms of $\ell_p$ neighborhoods.
At this point, we solve the adversarial problem \eqref{eq:low_rank_adversarial} in the framework of conditional gradient methods.
The analytical solution of LMO \eqref{eq:general_definition_LMO} for a nuclear ball of radius $\rho$ is given by
\BEQ\label{eq:LMO_nuclear}
\text{LMO}_{||\cdot||_{S1}\leq \rho}(M) \triangleq \rho~U_1 V_1^T,
\EEQ
where $U_1, V_1$ are the first columns of matrices $U$ and $V$ in the SVD decomposition of matrix $M$ given by $U S V^T$. For $q$-Schatten norm (with $q>1$) the LMO has also a closed form solution involving the full singular decomposition (see e.g., \cite[Lemma 7]{garber2015faster}). Solving LMO involves computing the right and left singular vectors $U_1$ and $V_1$ which are associated to the largest singular value $\rho$. Lanczos algorithm can be used to calculate singular vectors corresponding to largest singular value, where the solution is found using the Krylov subspace formed by the columns of matrix $M$. This demonstrates the computational efficiency of Frank-Wolfe methods as apposed to the other optimization approaches such as projected gradient descent, which requires the full SVD computation in each iteration. 
Qualitatively,  adversarial perturbations in nuclear norm add blurring effect to the original images, as for instance is depicted in Figure \ref{fig:nuclear_blur}. Thus, this can potentially pose a risk in some security scenarios, when such perturbations could be perceived as simple alterations of the image rather than a malware deformation of it, e.g., see \cite{gilmer2018motivating} for real-world scenarios. 

\subsection{Group constraints}\label{ssec:group_constraints}
In this section we demonstrate how to leverage weighted group norms in order to localize the low-rank perturbations. Group-norms are defined by a partition of the pixels' coordinates into groups. For instance, such a partition can be adapted from a segmentation of the sample image. These group-norms are a combination of two norms: a local one applied on vectors formed by each group of pixel values, and a global one applied on the vectors of the norms of all the groups. Here, we consider the nuclear norm as the local norm and the global $\ell_1$ norm to induce sparsity at the group level. Considering such norms provides some tools to substantially control the perturbations restricted to desirable parts to craft adversarialy perturbed images.

\paragraph{Nuclear Group Norm.}
Let $\mathcal{G}$ be an ensemble of groups of pixels' coordinates of the tensor image of $(c,h,w)$, where each element $g\in\mathcal{G}$ is a set of pixel coordinates'. Then for $x\in\mathbb{R}^{c\times h \times w}$ we define $\mathcal{G}$-nuclear group-norm as
\begin{equation}\label{eq:group_norm_schatten}
||x||_{\mathcal{G},1,p} = ||\,||x[g]||_{S(1), g\in\mathcal{G}}\,||_p,
\end{equation}
 with $p\in [1,\infty[\cup\{\infty\}$ (see for instance \cite{tomioka2013convex}.) When $\mathcal{G}$ is a partition of the pixels, $||\cdot||_{\mathcal{G},1,S(1)}$ is a norm. The nuclear group-norm allows to localize the blurring effect of the nuclear norm. Indeed, the LMO of $\mathcal{G}$-nuclear group-norm is given by
\begin{equation}
\label{eq:LMO_group_Schatten-1}
  \text{LMO}_{||\cdot||_{\mathcal{G}, 1, S1}\leq \rho}(M) \triangleq \left\{
    \begin{split}
    \rho ~ U^{(g)}_1  \big(V^{(g)}_1\big)^T\\
    0~\text{ otherwise}
    \end{split}
  \right.~,
\end{equation}
where $g^*=\underset{g\in\mathcal{G}}{\text{argmax }}\big|\big|M[g]\big|\big|_{S1}$ and the singular value decomposition of $M[g]$ for each group $g$ is given by $U^{(g)} S^{(g)} \big(V^{(g)}\big)^T$.
When solving \eqref{eq:opt_problem_adversarial} with such norms, each iteration of the conditional gradient will add to the adversarial perturbation a vertex of the form described by \eqref{eq:LMO_group_Schatten-1}, \textit{i.e.} a matrix of rank-one on the rectangle defined by the group of pixels in $g\in\mathcal{G}$. Note that the only modification for the approximate solution of group nuclear ball versus nuclear ball is the solution to LMO problem, and the rest of conditional gradient method for both of the distortion sets is similar. 

\paragraph{Different Distortion Radius per Group.}
When perturbing an image, modification in the pixel regions with high variance are typically harder to perceive than pixel modification in low variance regions. This knowledge was leveraged in \cite{luo2018towards} or in the $\sigma$-map of \cite[\S 2.2.]{croce2019sparse} to craft more imperceptible adversaries. Weighted nuclear group norms allow to search adversarial perturbations with different distortion radius across the image. With some $w_g>0$, the weighted nuclear group norm is defined as
\BEQ
||x||_{\mathcal{G}, 1, S(1), w} = \sum_{g\in\mathcal{G}}{ w_g ||x[g]||_{S(1)}},
\EEQ
and the LMO for weighted nuclear group-norm is then obtained as
\begin{equation}
  \begin{aligned}\label{eq:LMO_group_Schatten}
  &\text{LMO}_{||\cdot||_{\mathcal{G}, 1, S1}\leq \rho}(M) \triangleq\\
    ~&\begin{cases}
    \frac{\rho}{w_{g^*}} ~ U^{(g^*)}_1  \big(V^{(g^*)}_1\big)^T~&\text{ on group of pixels } g^*\\
    $0~$&\text{ otherwise}
  \end{cases}
  \end{aligned}
\end{equation}
where $g^*=\underset{g\in\mathcal{G}}{\text{argmax }}\frac{1}{w_g}\big|\big|M[g]\big|\big|_{S1}$ and the singular value decomposition of $M[g]$ for each group $g$ is given by $U^{(g)} S^{(g)} \big(V^{(g)}\big)^T$. In particular, this means that the solution corresponding to the group associated with $g$ have a nuclear radius of $\frac{\rho}{w_g}$ and the weights $w_g$ which allows to control the distortion in each group of pixels. The weights can be customized by the attacker to impose perturbation in desirable regions of the image. For instance, the weights can be chosen in inverse correlation with the variance of pixel regions to make the perturbations more targeted. 

\section{Structure Enhancing Algorithm for Adversarial Examples}\label{sec:FW_for_adversarial_examples}
We apply Frank-Wolfe algorithms \cite{frank1956algorithm}, a.k.a. conditional gradient algorithms \cite{levitin1966constrained}, for problem \eqref{eq:opt_problem_adversarial}. Given the conditional gradient optimization framework, the algorithm \ref{alg:vanilla_FW} can iteratively find the adversarial perturbation to fool the network.
For specific constraint structures such as the distortion set introduced earlier, conditional gradient algorithms naturally trades off between the convergence accuracy and the structured solutions in the early iterations.

\begin{algorithm}[h!]
   \caption{Vanilla Frank-Wolfe}
   \label{alg:vanilla_FW}
\begin{algorithmic}
   \STATE {\bfseries Input:} Original image $\xx_0$
    \FOR{$t=0,\cdots, T$}
        \STATE $s_{t} = \text{LMO}_{\mathcal{C}}\big(-\nabla \mathcal{L}(\xx_t)\big)$.
        \STATE $\gamma_t=\text{LineSearch}(x_{t}, s_t-x_t)$
        \STATE $x_{t+1} = (1 - \gamma_t)x_{t} + \gamma_t s_{t}$
   \ENDFOR
\end{algorithmic}
\end{algorithm}

For almost all the distortion sets which we consider in this work, LMO has a closed form solution. Note that the LMO has a low computational requirement as opposed to the projection based approaches. In particular, LMO requires only to compute the first singular vectors, while comparably projection steps demand the full SVD matrix to find the solution in each iteration. 
Provided the upper-bound for the Lipschitz constant $L$ of the adversarial loss is known, we apply the short step size $\gamma_t=\text{clip}_{[0,1]}(\langle -\nabla f(x_{t}), s_t-x_t\rangle/{L ||s_t - x_t||^2})$ for the optimization method. This is the only parameters that should be tuned in the algorithm, which makes the method more versatile to many models as compared to attacks which require hyperparameter tuning such as CW attacks \cite{carlini2017towards}.
\begin{table*}
\centering
\caption{MNIST and CIFAR-10 extensive white-box attack results. FWnucl 20$\,^{*}$: FWnucl with $\epsilon_{S1} = 1$. FWnucl 20$\,^{+}$: FWnucl with $\epsilon_{S1} = 3$. On MNIST (resp. CIFAR-10) PGD and FGSM have a total perturbation scale of 76.5/255 (0.3) (resp. 8/255 (0.031)), and step size 2.55/255 (0.01) (resp. 2/255 (0.01)). PGD runs for 20 iterations. We reproduce the ME-Net and Madry defense with same training hyper-parameters.}
\label{tab:attack_acc}
\begin{tabular}{@{\hspace{1cm}}ll ccccc}
\toprule
\multirow{2}{*}{\textbf{Network}} & \multirow{2}{*}{\textbf{Training}}  & \multirow{2}{*}{Clean} &
\multicolumn{4}{c}{\textbf{Accuracy under attack}}
\\\cmidrule{4-7}
 & {\textbf{Model}} &  & FWnucl 20$\,^{*}$ & FWnucl 20$\,^{+}$ & PGD 20 &  FGSM  \\
\midrule
\multicolumn{6}{l}{\textbf{MNIST}}\\\cmidrule{1-1}
\multirow{2}{*}{LeNET} & {Madry} & 98.38& {\bf 95.26} & {\bf 92.76} & 95.79 & 96.59 \\
& {ME-Net} & 99.24 & 97.63 & 75.41 & 74.88 & {\bf  46.18} \\
\addlinespace
 \midrule
 \multirow{2}{*}{SmallCNN} & {Madry} & 99.12 & 98.19 & 96.66 & {\bf 95.77} & 97.95 \\
& {ME-Net} & 99.42 & 89.56 & 78.65 & 76.84 & {\bf 54.09} \\
\addlinespace
 \midrule
\multicolumn{6}{l}{\textbf{CIFAR-10}}\\\cmidrule{1-1}
\multirow{2}{*}{ResNet-18} & {Madry} & 81.25 & {\bf 44.28} & {\bf 3.06} & 49.95 & 55.91 \\
& {ME-Net} & 93.45 & 29.66 & {\bf 4.01} & 4.99 & 44.80 \\
\addlinespace
\midrule
\multirow{2}{*}{WideResNet} & {Madry} & 85.1 & {\bf 43.16} & {\bf 2.82} & 52.49 & 59.06 \\
& {ME-Net} & 95.27 & 40.09 & 16.04 & {\bf 12.73} & 59.33 \\
\midrule
\multirow{2}{*}{ResNet-50} & {Madry} & 87.03 & {\bf 40.97} & {\bf 2.64} & 53.01 & 61.44  \\
& {ME-Net} &  92.09 & 47.66 & 17.81 & {\bf 9.14} & 58.51\\
\addlinespace
\bottomrule
\end{tabular}
\end{table*}

It is well-known that for non-convex, objective functions \textit{e.g.} the adversarial losses, injecting noise might be useful to escape from local optimums. This noise could be added either via random starts or via randomized block-coordinate methods. Under some additional conditions,  \cite{kerdreux2018frank} propose a version of Frank-Wolfe that solves linear minimization oracles on random subsets of the constraint sets. Here we consider subsampling the  image channels, i.e.,  \textit{i.e.} $||x||_{color, S1}= \sum_{c=1}^{3}||x_c||_{S1}$ where $x_c$ is one of the image channels. 
Note that we did not impose the box constraints which demonstrate that the values of image elements should be inside the interval $[0,1]^d$. To impose this restriction, we clamp the last iteration of optimization process to satisfy box constraints. Although this approach does not guarantee the convergence to a saddle point but removes the need to compute the LMO over the intersection of two sets, which is non-trivial. 

\section{Numerical Experiments}\label{sec:numerical_experiments}
This section aims at evaluating the adversarial accuracy of adversarial examples using Frank-Wolfe algorithms to the adversarial problem \eqref{eq:opt_problem_adversarial} with nuclear balls as distortion sets, which we refer as FWnucl. The complementary results for Frank-Wolfe with group norms and random initialization are provided in appendix.

\paragraph{Experiments Goal.} We tested FWnucl white-box attack against two baselines of defenses for untargeted attacks. The first is \cite{madry2017towards}, the state-of-the-art defense against white-box attacks. It uses the training images augmented with adversarial perturbations to train the network. The second one \cite{yang2019me} leverages matrix estimation techniques as a pre-processing step; each image is altered by randomly masking various proportions of the image pixels' and then reconstructed using matrix estimation by nuclear norm. For a given training image, this approach produces a group of images that are used during training, see \cite[\S 2.3.]{yang2019me} for more details. This provides a non-differentiable defense technique, \textit{i.e.} a method that cannot be straightforwardly optimized via back-propagation algorithms, and was reported to be robust against methods in \cite{athalye2018obfuscated} by circumventing the obfuscated gradients defenses. Qualitatively it leverages a structural difference between the low-rank structure of natural images and the adversarial perturbations that are not specifically designed to share the same structures.
We also evaluate our method against provably robust model trained with randomized smoothing \cite{cohen2019certified}. In the randomized smoothing, a provably robust classifier is derived from the convolution of base classifier with the isotropic Gaussian distribution of variance $\sigma^2$. This approach provides provable certified bounds in $L_2$ norm for the smoothed classifier. We show that the structured attacks can bring down the accuracy of model to the certified accuracy in almost all the smoothed models. Overall, a key motivation of our experiments is to propose adversarial examples with specific structures, serving at least as a sanity check for defense approaches. 
\paragraph{Experiment Settings.} We assess the accuracy of networks in different scenarios over MNIST and CIFAR-10 testsets. For ImageNet we randomly selected $4000$ from  the ImageNet validation set that are correctly classified.
For defense evaluation, for MNIST we use the LeNet model with two convolutional layers similar to \cite{madry2017towards} and SmallCNN with four convolutional layers followed by three fully connected layers as in \cite{carlini2017towards}. For CIFAR-10 dataset we use ResNet-18 and its wide version WideResNet and ResNet-50. For the ImageNet dataset we use ResNet-50 architecture. 

We report the adversarial accuracy of FWnucl along with those of classical attack methods like Fast Gradient Sign Method (FGSM) \cite{goodfellow2014explaining}, and Projected Gradient Descent (PGD) \cite{madry2017towards} to solve adversarial problem \eqref{eq:opt_problem_adversarial} using $\ell_{\infty}$ ball as the distortion set. FGSM generates adversarial examples with a single gradient step, while PGD is a more powerful adversary that performs a multi-step variant of FGSM. For each technique, in Table \ref{tab:attack_acc} we report accuracy as the percentage of adversarial examples that are classified correctly. We repeated the experiments several times to insure the results are general. These numerical experiments demonstrate that the attack success rates for FWnucl are comparable to the classical ones in an imperceptibility regime while also retaining specific structures in the perturbation. Note that FGSM for ME-Net provides better success rate (lower adversarial accuracy) compared to PGD which is the result of gradient masking generated by ME-Net over MNIST dataset. 

Table \ref{tab:attack_acc} also shows that FWnucl with $\epsilon = 3$ significantly performs better than other attacks. We attribute this difference to the fact that FWnucl has tendency to induce low-rank solutions, leading in global structures perturbation in images without any $\ell_p$ norm restrictions. This key characteristic of FWnucl makes it orthogonal to the existing adversarial attacks. FWnucl is specifically designed to iterate over solutions that lie on low-dimensional faces of the feasible set, as low-dimensional faces of the feasible region contain desirable well-structured low-rank matrices. 
\begin{table*}
\centering
\caption{ImageNet extensive white-box attack results on 4000 randomly selected images. FWnucl and PGD runs for 100 iterations. We reproduce Madry defense in ${\ell_{\infty}}$ norm with $\epsilon = 8/255$. In the table we report accuracy (Acc) versus the average of generated ${\ell_{2}}$ distortion for each attack.} 
\label{tab:attack_acc_imagenet}
\begin{tabular}{@{\hspace{0.2cm}}ll cc@{\hspace{0.05cm}}|@{\hspace{0.05cm}}cc@{\hspace{0.05cm}}|@{\hspace{0.05cm}}cc@{\hspace{0.05cm}}|@{\hspace{0.05cm}}cc@{\hspace{0.05cm}}|@{\hspace{0.05cm}}cc@{\hspace{0.05cm}}|@{\hspace{0.05cm}}cc@{\hspace{0.05cm}}|@{\hspace{0.05cm}}c}
\toprule
\multirow{3}{*}{\textbf{Network}} & {\textbf{Training}}  & \multirow{3}{*}{Clean} &
\multicolumn{6}{c}{FWnucl} & \multicolumn{6}{c}{PGD} \\\cmidrule{4-15}
& \multirow{2}{*}{\textbf{Model}} &  & \multicolumn{2}{c}{$\epsilon_{S1} = 1$} & \multicolumn{2}{c}{$\epsilon_{S1} = 3$} &  \multicolumn{2}{c}{$\epsilon_{S1} = 5$} & \multicolumn{2}{c}{$\epsilon = 2/255$} & \multicolumn{2}{c}{$\epsilon = 4/255$} & \multicolumn{2}{c}{$\epsilon = 8/255$}
\\\cmidrule{4-15}
 &  &  & Acc & $\ell_{2}$ & Acc & $\ell_{2}$ & Acc & $\ell_{2}$ &  Acc & $\ell_{2}$ & Acc & $\ell_{2}$ &  Acc & $\ell_{2}$ \\
\midrule
\multirow{2}{*}{ResNet-50} & {Standard} & 80.55& 19.67 & 0.69 & 1.62 & 1.27 & 0.17 & 1.68 & 0.2 & 2.53 & 0.0 & 4.55 &  0.0 & 8.53 \\
& {Madry} & 50.02& 38.3 & 1.45 & 16.8 & 3.82 & 6.62 & 5.80 & 42.07 & 2.97 & 34.52 & 5.90 &  18.9 & 11.69 \\
\bottomrule
\end{tabular}
\end{table*}
In Table \ref{tab:attack_acc_imagenet} we provided the adversarial accuracy for standardly and adversarialy trained models over ImageNet dataset. The results show that the attacks created by PGD show at least 50\% increase in $\ell_2$ norm distortion compared with FWnucl. Note that enlarging the radius of norm ball for PGD attack significantly increases the distortion while for FWnucl the increase in distortion rate is not fierce per increasing the nuclear ball radius. It confirms our earlier intuition that FWnucl is designed to selectively add distortion to pixels which are important for the label predictions. For adversarialy trained model, the robust accuracy for FWnucl is significantly lower that the counterparts from PGD. It indicates that FWnucl generate patterns that the robust models may not be robust to them.
\begin{table}
\setlength{\tabcolsep}{1pt}
\centering
\caption{Comparison of the white-box attacks for  CIFAR-10 on ResNet-18 adversarially trained. PGD, on the $\ell_{\infty}$ ball, and FGSM have a total perturbation scale of 8/255 (0.031), and step size 2/255(0.01).
FWnucl 20$\,^{*}$: FWnucl with $\epsilon_{S1} = 1$. FWnucl 20$\,^{+}$: FWnucl with $\epsilon_{S1} = 3$.
}
\label{tab:norms_cifar}
\begin{tabular}{@{}lc ccc@{}}
\toprule
&\multicolumn{2}{c}{\textbf{ResNet-18}}&\multicolumn{2}{c}{\textbf{ResNet-50}}
\\\cmidrule{2-5}
{{Attack}} & Mean $\ell_2$ & Mean $\norm{\cdot}_{{S1}}$ & Mean $\ell_2$ & Mean $\norm{\cdot}_{{S1}}$ \\
\midrule
{FWnucl 20$\,^{*}$} & {\bf 1.38} & {\bf 0.91}  & {\bf 1.31} & {\bf 0.91}  \\ 
{FWnucl 20$\,^{+}$} & 3.37 & 2.72 &3.00 & 2.65 \\
{PGD 20}& 1.68 & 3.88 & 1.66 &3.89\\
{FGSM}& 1.73 & 4.04 & 1.73 & 4.10 \\
\bottomrule
\end{tabular}
\end{table}
In Table \ref{tab:norms_cifar} we report the mean $\ell_2$, and nuclear norms of the adversarial noise over all attacks in Table \ref{tab:attack_acc} for the CIFAR-10 dataset (see the Appendix for MNIST dataset). Our method with $\epsilon_{S1} = 1$ generates perturbations with almost $7x$ and $10x$ lower $\ell_2$ norm for the MNIST dataset. Interestingly, the adversarial examples for FWnucl show significantly lower nuclear norm.

\begin{figure}
\subfloat{
\centering
\includegraphics[width=0.5\linewidth]{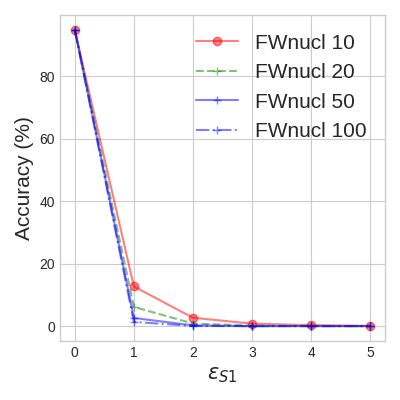}}%
\subfloat[]{
\centering
\includegraphics[width=0.5\linewidth]{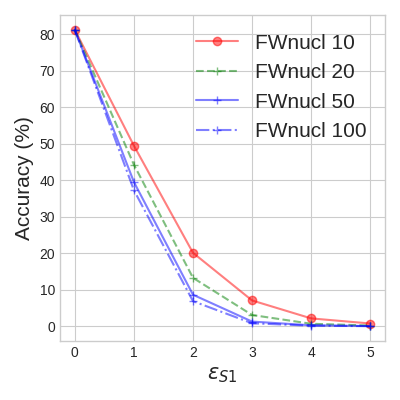}}%
\setlength{\abovecaptionskip}{2pt}
\caption{Accuracy of standard model (left) and robust model of Madry (right) on ResNet-18 for CIFAR-10, versus the nuclear ball radius when varying the number of steps.} 
\label{fig:attack_FW_radius_increase_steps}
\end{figure}

Figure \ref{fig:attack_FW_radius_increase_steps} summarizes the results for FWnucl with varying $\epsilon_{S1}$ for standard and robust model on CIFAR-10. The figure shows FWnucl algorithm noticeably drops the accuracy rate by increasing the radius $\epsilon_{S1}$. The performance of different FWnucl methods is slightly different, as the higher number of FWnucl steps may gain better performance.
\paragraph{Imperceptibility nuclear threshold.}

\begin{figure}
\subfloat[]{
\centering
\includegraphics[width=0.97\linewidth]{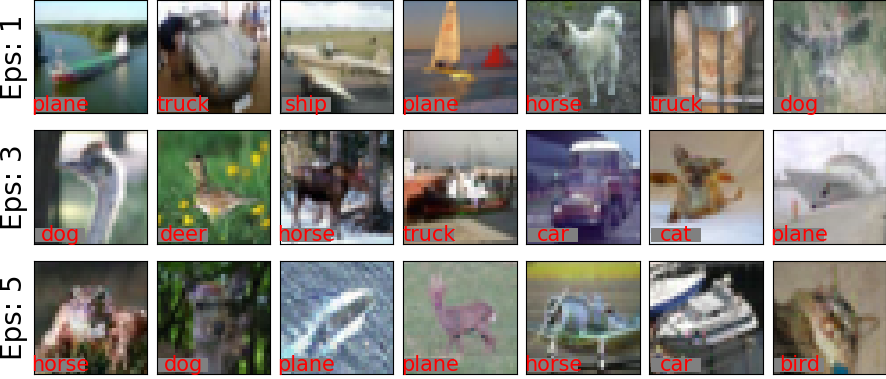}}%
\setlength{\abovecaptionskip}{2pt}
\caption{FWnucl adversarial examples for the CIFAR-10 dataset for different radii. The fooling label is shown on the image.}
\label{fig:mnist-cifar-images}
\end{figure}

We illustrate in Figure \ref{fig:mnist-cifar-images} some adversarial examples generated by FWnucl, for three different values of epsilon. The imperceptibility threshold exclusively depends on the dataset. On CIFAR-10, we qualitatively observed that with $\epsilon_{S1} = 1 $, all adversarial examples are perceptually identical to the original images. Also as the dataset becomes more complex, the tolerance of imperceptibility to nuclear ball radius values $\epsilon_{S1}$ increases; on ImageNet we realized the imperceptibility threshold is $\epsilon_{S1}=10$.

\begin{figure}
\centering
\subfloat[Basset]{{\includegraphics[scale=0.17]{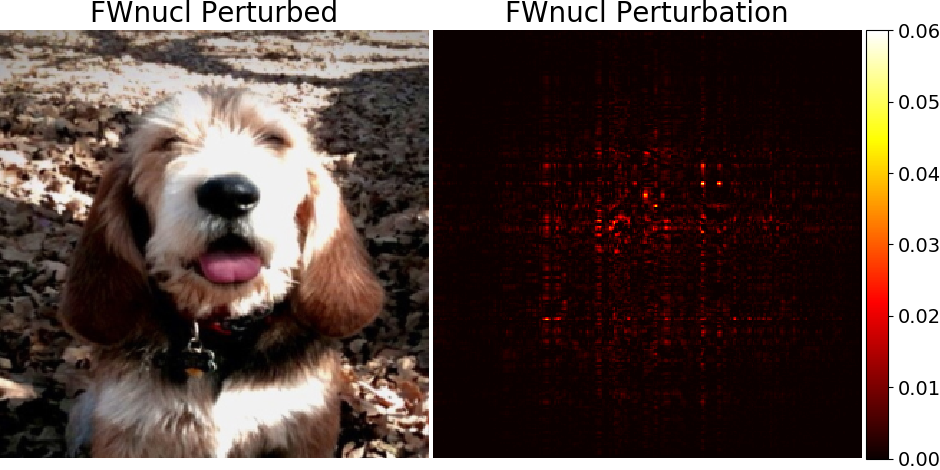}}}
\subfloat[Bouvier des Flandres]{{\includegraphics[scale=0.17]{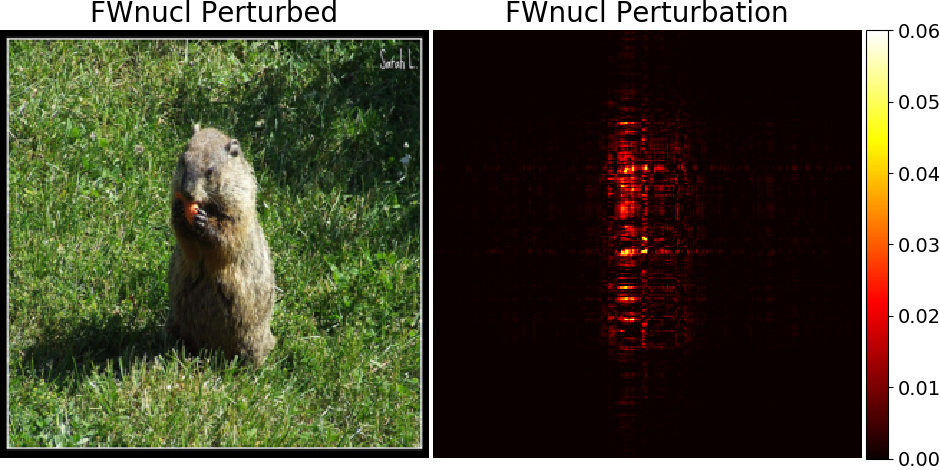}}}

\subfloat[Hog]{{\includegraphics[scale=0.17]{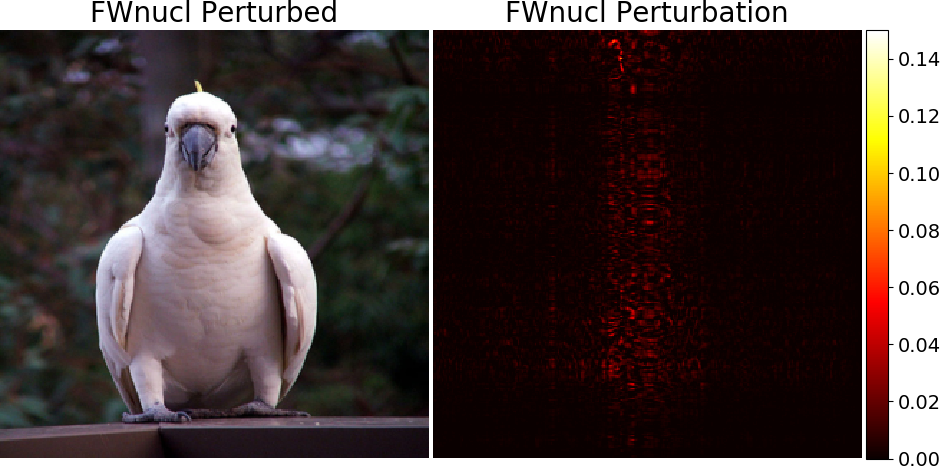}}}
\subfloat[Custard Apple]{{\includegraphics[scale=0.17]{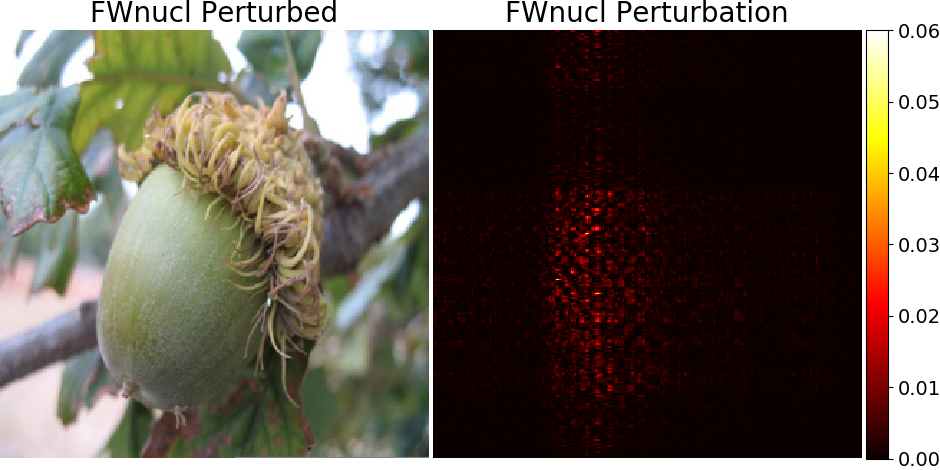}}}

\caption{The images display some structural pattern of FWnucl perturbations for the ImageNet dataset on DenseNet121 architecture for various level of distortion, standardly trained. Observe that the adversarial perturbed pixels are accumulated on the areas containing semantic information about the image. FWnucl is conducted with $\epsilon_{S1}=5$ and $20$ iterations.}\label{fig:distortion_imagenet}
\end{figure}
\begin{figure}
\centering
\subfloat[spoonbill]{{\includegraphics[scale=0.17]{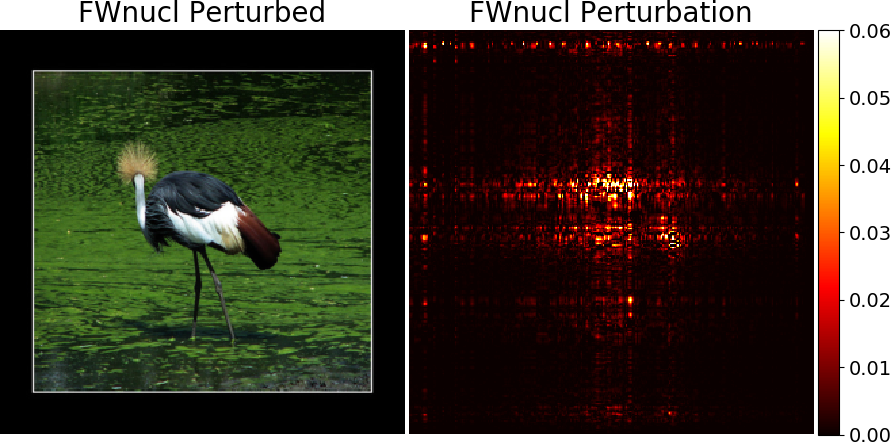}}}
\subfloat[spoonbill]{{\includegraphics[scale=0.17]{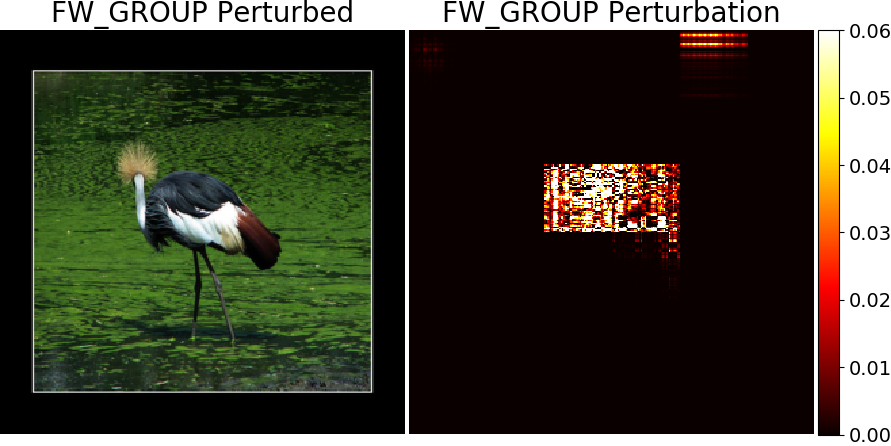}}}

\subfloat[pickup, pickup truck]{{\includegraphics[scale=0.17]{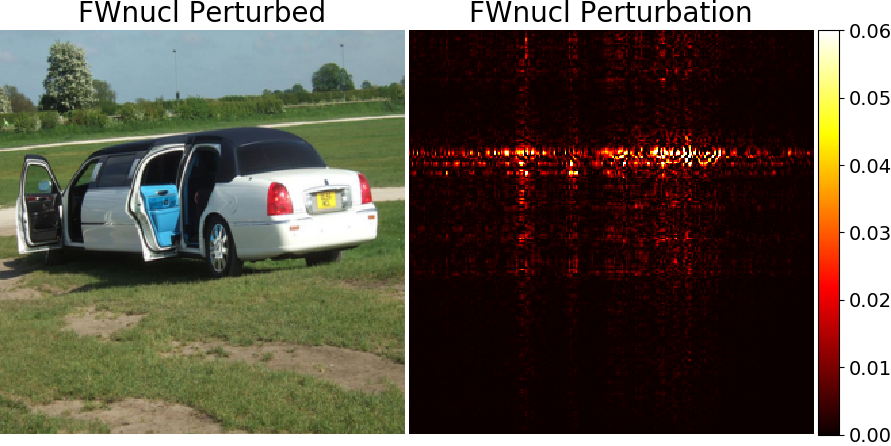}}}
\subfloat[racer, race car]{{\includegraphics[scale=0.17]{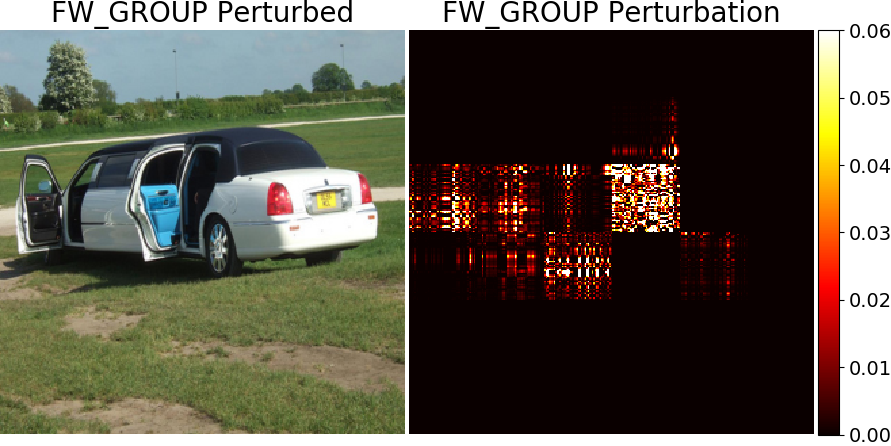}}}

\caption{The images display some structural pattern of FWnucl (left) versus Group-FWnucl (right) perturbations for the ImageNet dataset on DenseNet121 architecture for the nuclear ball of radius 5. Observe that the blurriness effect and perturbed pixels for images crafted by Group-FWnucl are localized and restricted to some specific groups of pixels.}\label{fig:distortion_imagenet-group}
\end{figure}

In Figure \ref{fig:distortion_imagenet}, we observe that the perturbations are particularly congregated around important regions (i.e., body, head), although there is not a universal configuration to detect specific features that are the  most  important  for  the  network. While the noise generated by PGD attack exhibits abrupt changes in pixel intensities (see Figure \ref{fig:example_diff_distortion}), the perturbation from FW has a continuous variations in pixel values. It is seen from the same figure that the conventional $\ell_p$ norm constrained methods e.g., FGSM, PGD do not encourage any structure and tends to generate perturbations even for pixels which might not be crucial for the label predictions, e.g., the background. However FWnucl only focuses on important regions of the image which might induce dramatic shift in the predictions.
 The FWnucl significantly reduces the number of perturbed elements in the image. For instance, the number of non-zero pixel coordinates for PGD and FGSM on ImageNet is respectively almost $11 x$ and $14 x$ larger than the number of non-zero pixel intensities for FWnucl with $\epsilon_{S1}=1$. 
In Figure \ref{fig:distortion_imagenet-group} we display the adversarial images and the corresponding perturbation generated by the nuclear norm versus group nuclear norm $||\cdot||_{\mathcal{G}, 1, S(1), w}$, where the weights $w$ are calculated based on local variance of each group. The figures show that Group-FWnul create perturbations which are more targeted and are localized to groups of pixels around the objective which are important for the classifier to make the prediction. 
It is important to characterize the type of deformation that arise with radii above the imperceptibility threshold as the imperceptibility regimes are not the only existing scenario for generating adversarial examples. In particular accuracy of robust networks quickly drop to zero in the regimes above the perceptibly regions, see Figure \ref{fig:attack_FW_radius_increase_steps} and appendix for adversarial accuracy with $\epsilon_{S1}$. In the nuclear ball case, as the radius $\epsilon_{S1}$ of the nuclear ball increases, the perturbation becomes perceptible with a blurring effect. Structure in the adversarial examples can be leveraged to create specific perceptible deformation effects that look natural to humans.
\paragraph{Provably robust models with certifiable bounds}
We also evaluate the performance of the proposed adversary against the robust models with provable certified bounds. Table \ref{tab:attack_acc_randomized_amoothing} shows that the accuracy of certified classifiers trained with randomized smoothing with standard deviation $\sigma = 0.5$. The results show that FWnucl can bring down the accuracy of the certifiably robust classifier up to the certified accuracy provided by provable defense methods for ResNet-18 and ResNet-50 models. It is also observed that there is a gap between the certified and adversarial accuracy for WideResNet model. 
\begin{table}
\centering
\caption{Evaluation of FWnucl against certifiably robust classifiers trained by randomized smoothing over CIFAR-10 dataset. Certified (accuracy) denotes the certifiable accuracy of smoothed classifier \cite{cohen2019certified}. Adversarial (accuracy) is the accuracy under FWnucl attack with the perturbation specified in the table. }
\label{tab:attack_acc_randomized_amoothing}
\begin{tabular}{l cccc}
\toprule
Certified & $l_2$ radius & 0.25 & 0.5 & 0.75\\
\cmidrule{1-5}
FWnucl & $l_{S_1}$ radius & 0.25 & 0.5 & 0.75  \\
 \midrule
 Certified & ResNet-18 & 52.26 & 39.94 & 27.49\\
 Adversarial & ResNet-18 & 48.66 & 38.37 & 28.28\\
 \midrule
 Certified & WideResNet & 55.9 & 28.97 & 16.83\\
 Adversarial & WideResNet & 53.89 & 41.65 & 30.75\\
 \midrule
 Certified & ResNet-50 &  50.04 & 37.11 & 24.77\\
 Adversarial &  ResNet-50 & 49.77 & 38.38 & 27.95\\
\addlinespace
\bottomrule
\end{tabular}
\end{table}
\section{Conclusion}
We consider adversarial attacks beyond  $\ell_p$ distortion set. Our proposed structured attacks  allow an attacker to design imperceptible adversarial examples with specific characteristics, like localized blurriness. Furthermore, in the imperceptible regime, some defensive techniques may rely on a lack of certain structured patterns in the adversarial perturbations. Evaluating robustness against various structured adversarial examples then seems to be a reasonable defense sanity check. Our method is a competitor to the methods designed to craft sparse and targeted perturbations while maintaining success rates similar to powerful attacks like PGD. 

{\small
\bibliographystyle{ieee_fullname}
\bibliography{main}
}

\end{document}

%% file: defs.tex
\def\xx{{\boldsymbol x}}

\newcommand{\BEAS}{\begin{eqnarray*}}
\newcommand{\EEAS}{\end{eqnarray*}}
\newcommand{\BEA}{\begin{eqnarray}}
\newcommand{\EEA}{\end{eqnarray}}
\newcommand{\BEQ}{\begin{equation}}
\newcommand{\EEQ}{\end{equation}}
\newcommand{\BIT}{\begin{itemize}}
\newcommand{\EIT}{\end{itemize}}
\newcommand{\BNUM}{\begin{enumerate}}
\newcommand{\ENUM}{\end{enumerate}}

\newcommand{\BA}{\begin{array}}
\newcommand{\EA}{\end{array}}

\newcommand{\argmin}{\mathop{\rm argmin}}

\newcommand{\norm}[1]{\left\lVert#1\right\rVert}

%% file: main.bbl
\begin{thebibliography}{10}\itemsep=-1pt

\bibitem{allen2017linear}
Zeyuan Allen-Zhu, Elad Hazan, Wei Hu, and Yuanzhi Li.
\newblock Linear convergence of a frank-wolfe type algorithm over trace-norm
  balls.
\newblock In {\em Advances in Neural Information Processing Systems}, pages
  6191--6200, 2017.

\bibitem{athalye2018obfuscated}
Anish Athalye, Nicholas Carlini, and David Wagner.
\newblock Obfuscated gradients give a false sense of security: Circumventing
  defenses to adversarial examples.
\newblock {\em arXiv preprint arXiv:1802.00420}, 2018.

\bibitem{candes2009exact}
Emmanuel~J Cand{\`e}s and Benjamin Recht.
\newblock Exact matrix completion via convex optimization.
\newblock {\em Foundations of Computational mathematics}, 9(6):717, 2009.

\bibitem{carlini2019evaluating}
Nicholas Carlini, Anish Athalye, Nicolas Papernot, Wieland Brendel, Jonas
  Rauber, Dimitris Tsipras, Ian Goodfellow, Aleksander Madry, and Alexey
  Kurakin.
\newblock On evaluating adversarial robustness.
\newblock {\em arXiv preprint arXiv:1902.06705}, 2019.

\bibitem{carlini2017towards}
Nicholas Carlini and David Wagner.
\newblock Towards evaluating the robustness of neural networks.
\newblock In {\em 2017 IEEE Symposium on Security and Privacy (SP)}, pages
  39--57. IEEE, 2017.

\bibitem{chen2018frank}
Jinghui Chen, Jinfeng Yi, and Quanquan Gu.
\newblock A frank-wolfe framework for efficient and effective adversarial
  attacks.
\newblock {\em arXiv preprint arXiv:1811.10828}, 2018.

\bibitem{cheung2017projection}
Edward Cheung and Yuying Li.
\newblock Projection free rank-drop steps.
\newblock {\em arXiv preprint arXiv:1704.04285}, 2017.

\bibitem{cohen2019certified}
Jeremy~M Cohen, Elan Rosenfeld, and J~Zico Kolter.
\newblock Certified adversarial robustness via randomized smoothing.
\newblock {\em arXiv preprint arXiv:1902.02918}, 2019.

\bibitem{croce2019sparse}
Francesco Croce and Matthias Hein.
\newblock Sparse and imperceivable adversarial attacks.
\newblock In {\em Proceedings of the IEEE International Conference on Computer
  Vision}, pages 4724--4732, 2019.

\bibitem{demyanov1970}
V.~F. Demyanov and A.~M. Rubinov.
\newblock Approximate methods in optimization problems.
\newblock {\em Modern Analytic and Computational Methods in Science and
  Mathematics}, 1970.

\bibitem{dudik2012lifted}
Miroslav Dudik, Zaid Harchaoui, and J{\'e}r{\^o}me Malick.
\newblock Lifted coordinate descent for learning with trace-norm
  regularization.
\newblock In {\em Artificial Intelligence and Statistics}, pages 327--336,
  2012.

\bibitem{dunn1979rates}
Joseph~C Dunn.
\newblock Rates of convergence for conditional gradient algorithms near
  singular and nonsingular extremals.
\newblock {\em SIAM Journal on Control and Optimization}, 17(2):187--211, 1979.

\bibitem{engstrom2017rotation}
Logan Engstrom, Brandon Tran, Dimitris Tsipras, Ludwig Schmidt, and Aleksander
  Madry.
\newblock A rotation and a translation suffice: Fooling cnns with simple
  transformations.
\newblock {\em arXiv preprint arXiv:1712.02779}, 2017.

\bibitem{fazel2001rank}
Maryam Fazel, Haitham Hindi, Stephen~P Boyd, et~al.
\newblock A rank minimization heuristic with application to minimum order
  system approximation.
\newblock Citeseer, 2001.

\bibitem{frank1956algorithm}
Marguerite Frank and Philip Wolfe.
\newblock An algorithm for quadratic programming.
\newblock {\em Naval research logistics quarterly}, 3(1-2):95--110, 1956.

\bibitem{freund2017extended}
Robert~M Freund, Paul Grigas, and Rahul Mazumder.
\newblock An extended frank--wolfe method with “in-face” directions, and
  its application to low-rank matrix completion.
\newblock {\em SIAM Journal on Optimization}, 27(1):319--346, 2017.

\bibitem{garber2013linearly}
Dan Garber and Elad Hazan.
\newblock A linearly convergent conditional gradient algorithm with
  applications to online and stochastic optimization.
\newblock {\em arXiv preprint arXiv:1301.4666}, 2013.

\bibitem{garber2013playing}
Dan Garber and Elad Hazan.
\newblock Playing non-linear games with linear oracles.
\newblock In {\em 2013 IEEE 54th Annual Symposium on Foundations of Computer
  Science}, pages 420--428. IEEE, 2013.

\bibitem{garber2015faster}
Dan Garber and Elad Hazan.
\newblock Faster rates for the frank-wolfe method over strongly-convex sets.
\newblock In {\em 32nd International Conference on Machine Learning, ICML
  2015}, 2015.

\bibitem{garber2018fast}
Dan Garber, Shoham Sabach, and Atara Kaplan.
\newblock Fast generalized conditional gradient method with applications to
  matrix recovery problems.
\newblock {\em arXiv preprint arXiv:1802.05581}, 2018.

\bibitem{gatys2017controlling}
Leon~A Gatys, Alexander~S Ecker, Matthias Bethge, Aaron Hertzmann, and Eli
  Shechtman.
\newblock Controlling perceptual factors in neural style transfer.
\newblock In {\em Proceedings of the IEEE Conference on Computer Vision and
  Pattern Recognition}, pages 3985--3993, 2017.

\bibitem{gilmer2018motivating}
Justin Gilmer, Ryan~P Adams, Ian Goodfellow, David Andersen, and George~E Dahl.
\newblock Motivating the rules of the game for adversarial example research.
\newblock {\em arXiv preprint arXiv:1807.06732}, 2018.

\bibitem{goodfellow2014explaining}
Ian~J Goodfellow, Jonathon Shlens, and Christian Szegedy.
\newblock Explaining and harnessing adversarial examples.
\newblock 2015.

\bibitem{guelat1986some}
Jacques Gu{\'e}lat and Patrice Marcotte.
\newblock Some comments on {Wolfe}'s ‘away step’.
\newblock {\em Mathematical Programming}, 1986.

\bibitem{guo2018low}
Chuan Guo, Jared~S Frank, and Kilian~Q Weinberger.
\newblock Low frequency adversarial perturbation.
\newblock {\em arXiv preprint arXiv:1809.08758}, 2018.

\bibitem{harchaoui2012large}
Zaid Harchaoui, Matthijs Douze, Mattis Paulin, Miroslav Dudik, and
  J{\'e}r{\^o}me Malick.
\newblock Large-scale image classification with trace-norm regularization.
\newblock In {\em 2012 IEEE Conference on Computer Vision and Pattern
  Recognition}, pages 3386--3393. IEEE, 2012.

\bibitem{jaggi2013revisiting}
Martin Jaggi.
\newblock Revisiting frank-wolfe: Projection-free sparse convex optimization.
\newblock In {\em Proceedings of the 30th international conference on machine
  learning}, number CONF, pages 427--435, 2013.

\bibitem{jaggi2010simple}
Martin Jaggi and Marek Sulovsk{\`y}.
\newblock A simple algorithm for nuclear norm regularized problems.
\newblock 2010.

\bibitem{kerdreux2020uniformly}
Thomas Kerdreux and Alexandre d'Aspremont.
\newblock Frank-wolfe on uniformly convex sets.
\newblock 2020.

\bibitem{kerdreux2018frank}
Thomas Kerdreux, Fabian Pedregosa, and Alexandre d'Aspremont.
\newblock Frank-wolfe with subsampling oracle.
\newblock {\em arXiv preprint arXiv:1803.07348}, 2018.

\bibitem{keskar2016large}
Nitish~Shirish Keskar, Dheevatsa Mudigere, Jorge Nocedal, Mikhail Smelyanskiy,
  and Ping Tak~Peter Tang.
\newblock On large-batch training for deep learning: Generalization gap and
  sharp minima.
\newblock {\em arXiv preprint arXiv:1609.04836}, 2016.

\bibitem{kurakin2016adversarial}
Alexey Kurakin, Ian Goodfellow, and Samy Bengio.
\newblock Adversarial examples in the physical world.
\newblock {\em arXiv preprint arXiv:1607.02533}, 2016.

\bibitem{lacoste2013affine}
Simon Lacoste-Julien and Martin Jaggi.
\newblock An affine invariant linear convergence analysis for frank-wolfe
  algorithms.
\newblock {\em arXiv preprint arXiv:1312.7864}, 2013.

\bibitem{FW-converge2015}
Simon Lacoste-Julien and Martin Jaggi.
\newblock On the global linear convergence of {F}rank--{W}olfe optimization
  variants.
\newblock In C. Cortes, N.~D. Lawrence, D.~D. Lee, M. Sugiyama, and R. Garnett,
  editors, {\em Advances in Neural Information Processing Systems}, volume~28,
  pages 496--504. Curran Associates, Inc., 2015.

\bibitem{langeberg2019effect}
Peter Langeberg, Emilio~Rafael Balda, Arash Behboodi, and Rudolf Mathar.
\newblock On the effect of low-rank weights on adversarial robustness of neural
  networks.
\newblock {\em arXiv preprint arXiv:1901.10371}, 2019.

\bibitem{lee2010practical}
Jason~D Lee, Ben Recht, Nathan Srebro, Joel Tropp, and Russ~R Salakhutdinov.
\newblock Practical large-scale optimization for max-norm regularization.
\newblock In {\em Advances in neural information processing systems}, pages
  1297--1305, 2010.

\bibitem{levitin1966constrained}
Evgeny~S Levitin and Boris~T Polyak.
\newblock Constrained minimization methods.
\newblock {\em USSR Computational mathematics and mathematical physics},
  6(5):1--50, 1966.

\bibitem{liu2018beyond}
Hsueh-Ti~Derek Liu, Michael Tao, Chun-Liang Li, Derek Nowrouzezahrai, and Alec
  Jacobson.
\newblock Beyond pixel norm-balls: Parametric adversaries using an analytically
  differentiable renderer.
\newblock 2018.

\bibitem{lu2017decoder}
Ming Lu, Hao Zhao, Anbang Yao, Feng Xu, Yurong Chen, and Li Zhang.
\newblock Decoder network over lightweight reconstructed feature for fast
  semantic style transfer.
\newblock In {\em Proceedings of the IEEE International Conference on Computer
  Vision}, pages 2469--2477, 2017.

\bibitem{luo2018towards}
Bo Luo, Yannan Liu, Lingxiao Wei, and Qiang Xu.
\newblock Towards imperceptible and robust adversarial example attacks against
  neural networks.
\newblock In {\em Thirty-Second AAAI Conference on Artificial Intelligence},
  2018.

\bibitem{madry2017towards}
Aleksander Madry, Aleksandar Makelov, Ludwig Schmidt, Dimitris Tsipras, and
  Adrian Vladu.
\newblock Towards deep learning models resistant to adversarial attacks.
\newblock {\em arXiv preprint arXiv:1706.06083}, 2017.

\bibitem{raghunathan2018certified}
Aditi Raghunathan, Jacob Steinhardt, and Percy Liang.
\newblock Certified defenses against adversarial examples.
\newblock {\em arXiv preprint arXiv:1801.09344}, 2018.

\bibitem{rauber2017foolbox}
Jonas Rauber, Wieland Brendel, and Matthias Bethge.
\newblock Foolbox: A python toolbox to benchmark the robustness of machine
  learning models.
\newblock In {\em Reliable Machine Learning in the Wild Workshop, 34th
  International Conference on Machine Learning}, 2017.

\bibitem{reed2016learning}
Scott~E Reed, Zeynep Akata, Santosh Mohan, Samuel Tenka, Bernt Schiele, and
  Honglak Lee.
\newblock Learning what and where to draw.
\newblock In {\em Advances in neural information processing systems}, pages
  217--225, 2016.

\bibitem{risser2017stable}
Eric Risser, Pierre Wilmot, and Connelly Barnes.
\newblock Stable and controllable neural texture synthesis and style transfer
  using histogram losses.
\newblock {\em arXiv preprint arXiv:1701.08893}, 2017.

\bibitem{schmidt2018adversarially}
Ludwig Schmidt, Shibani Santurkar, Dimitris Tsipras, Kunal Talwar, and
  Aleksander Madry.
\newblock Adversarially robust generalization requires more data.
\newblock In {\em Advances in Neural Information Processing Systems}, pages
  5014--5026, 2018.

\bibitem{sen2019should}
Ayon Sen, Xiaojin Zhu, Liam Marshall, and Robert Nowak.
\newblock Should adversarial attacks use pixel p-norm?
\newblock {\em arXiv preprint arXiv:1906.02439}, 2019.

\bibitem{shalev2011large}
Shai Shalev-Shwartz, Alon Gonen, and Ohad Shamir.
\newblock Large-scale convex minimization with a low-rank constraint.
\newblock {\em arXiv preprint arXiv:1106.1622}, 2011.

\bibitem{sharif2018suitability}
Mahmood Sharif, Lujo Bauer, and Michael~K Reiter.
\newblock On the suitability of lp-norms for creating and preventing
  adversarial examples.
\newblock In {\em Proceedings of the IEEE Conference on Computer Vision and
  Pattern Recognition Workshops}, pages 1605--1613, 2018.

\bibitem{stutz2019disentangling}
David Stutz, Matthias Hein, and Bernt Schiele.
\newblock Disentangling adversarial robustness and generalization.
\newblock In {\em Proceedings of the IEEE Conference on Computer Vision and
  Pattern Recognition}, pages 6976--6987, 2019.

\bibitem{tomioka2013convex}
Ryota Tomioka and Taiji Suzuki.
\newblock Convex tensor decomposition via structured schatten norm
  regularization.
\newblock In {\em Advances in neural information processing systems}, pages
  1331--1339, 2013.

\bibitem{wong2017provable}
Eric Wong and J~Zico Kolter.
\newblock Provable defenses against adversarial examples via the convex outer
  adversarial polytope.
\newblock {\em arXiv preprint arXiv:1711.00851}, 2017.

\bibitem{wong2020learning}
Eric Wong and J~Zico Kolter.
\newblock Learning perturbation sets for robust machine learning.
\newblock {\em arXiv preprint arXiv:2007.08450}, 2020.

\bibitem{wong2019wasserstein}
Eric Wong, Frank~R Schmidt, and J~Zico Kolter.
\newblock Wasserstein adversarial examples via projected sinkhorn iterations.
\newblock {\em arXiv preprint arXiv:1902.07906}, 2019.

\bibitem{xu2018structured}
Kaidi Xu, Sijia Liu, Pu Zhao, Pin-Yu Chen, Huan Zhang, Quanfu Fan, Deniz
  Erdogmus, Yanzhi Wang, and Xue Lin.
\newblock Structured adversarial attack: Towards general implementation and
  better interpretability.
\newblock {\em arXiv preprint arXiv:1808.01664}, 2018.

\bibitem{Subspace19}
Ziang Yan, Yiwen Guo, and Changshui Zhang.
\newblock Subspace attack: Exploiting promising subspaces for query-efficient
  black-box attacks, 2019.

\bibitem{yang2020randomized}
Greg Yang, Tony Duan, Edward Hu, Hadi Salman, Ilya Razenshteyn, and Jerry Li.
\newblock Randomized smoothing of all shapes and sizes.
\newblock {\em arXiv preprint arXiv:2002.08118}, 2020.

\bibitem{yang2019me}
Yuzhe Yang, Guo Zhang, Dina Katabi, and Zhi Xu.
\newblock Me-net: Towards effective adversarial robustness with matrix
  estimation.
\newblock {\em arXiv preprint arXiv:1905.11971}, 2019.

\bibitem{zhang2019towards}
Huan Zhang, Hongge Chen, Chaowei Xiao, Sven Gowal, Robert Stanforth, Bo Li,
  Duane Boning, and Cho-Jui Hsieh.
\newblock Towards stable and efficient training of verifiably robust neural
  networks.
\newblock {\em arXiv preprint arXiv:1906.06316}, 2019.

\end{thebibliography}
